\documentclass[10pt,twocolumn,letterpaper,hyphens]{article}

\usepackage{bm}
\usepackage{multirow}
\usepackage{graphicx}
\usepackage{makecell}
\usepackage{diagbox}
\usepackage{fontawesome5}  
\usepackage{pifont} 
\usepackage[ruled,vlined]{algorithm2e}
\usepackage{hyperref}

\usepackage[pagenumbers]{cvpr} 

\definecolor{cvprblue}{rgb}{0.21,0.49,0.74}

\def\paperID{} 
\def\confName{CVPR}
\def\confYear{2026}

\title{MedKCO: Medical Vision-Language Pretraining via Knowledge-Driven Cognitive Orchestration}

\author{
Chenran Zhang$^{1,2}$, Ruiqi Wu$^{1,2}$, Tao Zhou$^{3}$, Yi Zhou$^{1,2}$ \thanks{Corresponding author is Yi Zhou.}\\
$^{1}$School of Computer Science and Engineering, Southeast University, China\\
$^{2}$Key Laboratory of New Generation Artificial Intelligence Technology and Its Interdisciplinary\\ Applications, Ministry of Education, China\\
$^{3}$School of Computer Science and Engineering, Nanjing University of Science and Technology, China\\
{\tt\small \{chenranzhang, ruiqiwu\}@seu.edu.cn}\quad{\tt\small \{taozhou.ai, yizhou.szcn\}@gmail.com}
}

\begin{document}
\maketitle
\begin{abstract}
Medical vision-language pretraining (VLP) models have recently been investigated for their generalization to diverse downstream tasks. However, current medical VLP methods typically force the model to learn simple and complex concepts simultaneously. This anti-cognitive process leads to suboptimal feature representations, especially under distribution shift. To address this limitation, we propose a \textbf{K}nowledge-driven \textbf{C}ognitive \textbf{O}rchestration for \textbf{Med}ical VLP (MedKCO) that involves both the ordering of the pretraining data and the learning objective of vision-language contrast. Specifically, we design a two level curriculum by incorporating diagnostic sensitivity and intra-class sample representativeness for the ordering of the pretraining data. Moreover, considering the inter-class similarity of medical images, we introduce a self-paced asymmetric contrastive loss to dynamically adjust the participation of the pretraining objective. We evaluate the proposed pretraining method on three medical imaging scenarios in multiple vision-language downstream tasks, and compare it with several curriculum learning methods. Extensive experiments show that our method significantly surpasses all baselines. \url{https://github.com/Mr-Talon/MedKCO.}
\end{abstract}
\section{Introduction}
\label{sec:intro}
Medical VLP aims to align medical images with their corresponding descriptions\cite{flair, keepfit, medclip, cxrclip}. However, the inherent characteristics of medical data make this alignment particularly challenging: 1) diagnostic difficulty varies across diseases, 2) sample representativeness within the same disease differs considerably, 3) medical images appear high inter-class similarity, whereas textual descriptions are discriminative. Simple random shuffling pretraining\cite{flair, keepfit, medclip, cxrclip} compels models to learn a heterogeneous high-dimensional alignment between medical images and texts before establishing fundamental anatomical and lesion concepts, overlooking the influence of cognitive orchestration during pretraining. This limits representation effectiveness and clinical applicability.

To address these challenges, we propose a curriculum-based \cite{curriculum, CLsurvey, neighbor, clipvg}, knowledge-driven cognitive orchestration for medical VLP, named MedKCO. Inspired by the ``zone of proximal development" \cite{mind, role, vygotsky2012thought, wass2014sharpening} in cognitive science, which emphasizes learning through gradually increasing complexity within the learner’s capacity, our method orchestrates pretraining from simple to complex concepts progressively based on medical domain knowledge. MedKCO includes a hierarchical curriculum for the order of the pretraining data and a self-paced asymmetric contrastive loss for the pretraining objective function.

\begin{figure*}[t]
\includegraphics[width=\textwidth]{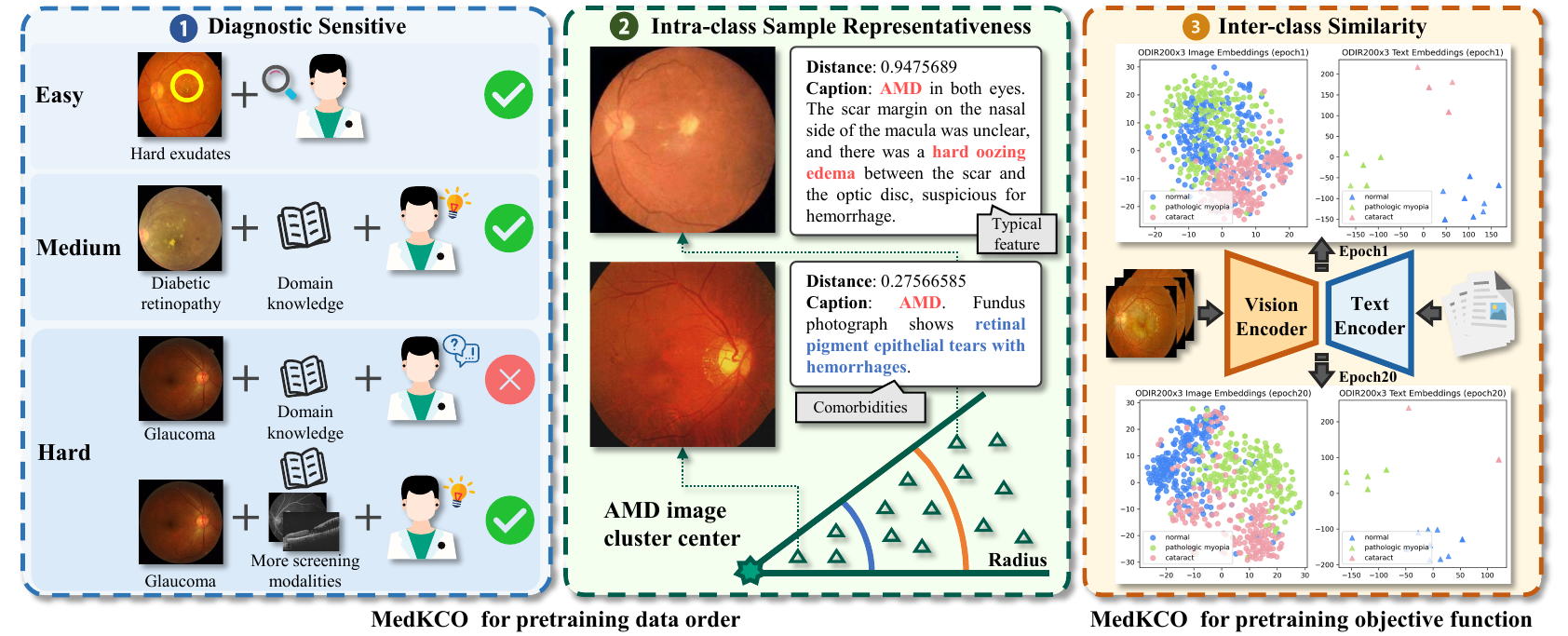}
\centering
\caption{Motivations of knowledge-driven cognitive orchestration. (a) Diagnostic sensitivity varies among different diseases. (b) Representativeness of intra-class samples exhibits variation. (c) High inter-class similarity in medical images at the beginning of the pretraining.} \label{fig1}
\end{figure*}

Specifically, previous works \cite{flair, keepfit,yao2025continual, medclip, cxrclip} always incorporate categorical label-supervised data with text-supervised data using fixed prompt templates. For the order of the pretraining data in cognitive orchestration, we design a \textbf{hierarchical curriculum} which divides the data into label-level and description-level, defined by the granularity of supervisory signals. 1) For \textbf{label-level} data, we find that the sensitivity of a single modality to detect the corresponding label varies considerably. As the color fundus photography example shown in \cref{fig1}(a), morphological signs such as “hard exudate” are directly observable, while the diagnosis of “diabetic retinopathy” and “glaucoma" requires deeper domain knowledge or more complementary modalities. Guided by this hierarchy of diagnostic sensitivity, we divide the label-level data into three stages of increasing cognitive difficulty, as shown in \cref{fig2}(left). 2) For \textbf{description-level} data, the representativeness of intra-class samples exhibits substantial variation attributed to individual variability and comorbidities, as shown in \cref{fig1}(b). Highly representative samples convey clearer disease concepts and facilitate early cognitive formation, whereas atypical cases demand more complex learning. Consequently, we partition the description-level data into multiple stages ordered by decreasing representativeness, as shown in \cref{fig2}(right).

When it comes to the objective function of pretraining in cognitive orchestration, an additional limitation arises from the high inter-class similarity of medical images. Standard symmetric contrastive loss often leads to an overly compact visual feature space in early pretraining, as shown in \cref{fig1}(c). However, medical textual descriptions typically exhibit clearer semantic distinctions. This asymmetry makes text-to-image alignment more challenging than image-to-text alignment, especially during the early pretraining stage. To alleviate this imbalance, we propose a \textbf{self-paced asymmetric contrastive loss}, which gradually increases the participation of text-to-image contrast, allowing the model to initially focus on simple alignment tasks and then progressively handle more complex alignment scenarios, analogous to the human cognitive process.

We introduce MedKCO, a knowledge-driven cognitive orchestration for medical VLP, which involves both the ordering of the pretraining data and the objective function of pretraining. We highlight our main contributions as follows:
\textbf{1)} For the ordering of pretraining data in the cognitive orchestration, we design a hierarchical curriculum based on the sensitivity of each modality to detect specific disease and the representativeness of intra-class samples.
\textbf{2)} For the pretraining objective of vision-language contrast, we develop a self-paced asymmetric contrastive loss that dynamically adapts the participation of different proxy tasks during pretraining.
\textbf{3)} We evaluated the proposed MedKCO pretraining method in three different medical imaging scenarios in various downstream datasets. Extensive experiments show that the proposed pretraining method achieves significant improvements in vision-language multimodal tasks.





\section{Related Works}
\subsection{Vision-Language Pretraining}
VLP models, such as CLIP \cite{clip}, ALIGN \cite{align}, FILIP\cite{filip}, leverage large-scale paired image-text data for contrastive learning, producing general representations transferable to downstream tasks. The objective of VLP is to maximize similarity for paired image-text and minimize unpaired ones in a multimodal space. In the biomedical domain, models such as MedCLIP \cite{medclip}, CXR-CLIP\cite{cxrclip}, FLAIR\cite{flair}, KeepFIT \cite{keepfit}, PLIP \cite{pathology}, CPLIP\cite{cplip}, PanDerm\cite{dermatology}, Derm1M\cite{derm1m} adapt this paradigm to radiology, fundus imaging, pathology, and dermatology. Most of these methods incorporate both categorical label-supervised data and text-supervised data to expand the pretraining corpus. However, these methods typically feed the training data to the model in random order, neglecting the influence of cognitive orchestration during the pretraining process. 

\begin{figure*}[t]
\includegraphics[width=\textwidth]{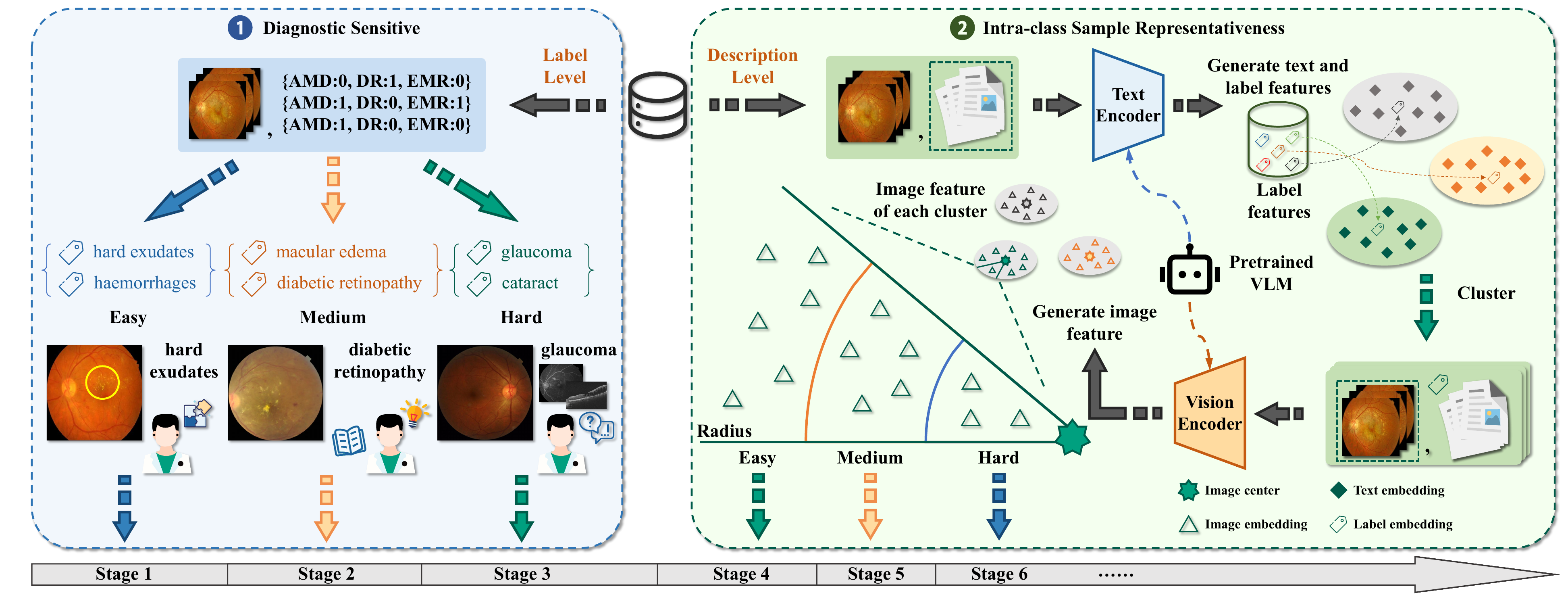}
\centering
\caption{Overview of the two-level curriculum. The pretraining data is divided into two distinct levels, label-level (left) and description-level (right). The label-level data is categorized into three stages according to the sensitivity of each modality to detect specific disease. The description-level data is clustered into the most relevant categories based on their textual descriptions. Subsequently, within each category, samples are divided into multiple stages according to the representativeness of their image features.} \label{fig2}
\end{figure*}

\subsection{Curriculum Learning}
Curriculum Learning (CL), introduced by Bengio \etal \cite{curriculum}, trains models progressively from simple to complex tasks, analogous to how humans acquire knowledge through structured curricula \cite{SPCL, CLsurvey}. When it comes to VLP, Srinivasan \etal \cite{cvprws} structures its curriculum based on the granularity of textual supervision, pretraining models in stages from object-level to instance-level understanding. Chen \etal \cite{vistruct} organizes its pretraining according to visual task difficulty, guiding the model through concept recognition, object grounding, attribute comprehension, relation detection, and event understanding. Despite these advances, the application of curriculum learning in medical VLP remains understudied. No existing work explicitly addresses the unique difficulty structure inherent in medical image–text data. Driven by domain knowledge in medical images, this work proposes a cognitive orchestration for pretraining from two perspectives: the ordering of pretraining data and the vision–language pretraining objective, building on the sensitivity of specific medical imaging modalities, intra-class variability in sample representativeness, and inter-class feature similarity.
\section{Methodology}
\subsection{Vision-Language Pretraining Framework \label{vlp}}
VLP leverages paired image–text data and vision–language contrastive learning to align visual and textual feature spaces, producing transferable representations for downstream tasks\cite{clip,align}. Our proposed pretraining method is model-agnostic, universally applicable to all medical VLP models. We denote the paired image–text data by $D_i=(x_i, y_i)$, the vision and text encoders by $E_v, E_t$, respectively. For label-level data, $y_i$ is generated from a domain knowledge template \cite{flair, medclip, cxrclip}, which first randomly selects a description from the domain knowledge dictionary to replace the label and then inserts the replaced label into the template to form the caption. For description-level data, $y_i$ is the original caption associated with the image.

When it comes to vision-language contrastive learning, the in-batch image to text and text to image similarities $S_{i,j}^{i2t}$ and $S_{i,j}^{t2i}$ are calculated through the encoded vision and language features after projection:
\begin{equation}
    S_{i,j}^{i2t}=sim(v_i, t_j),\quad S_{i,j}^{t2i}=sim(t_i, v_j),
\label{eq1}
\end{equation}
where $sim(\cdot)$ is any similarity function, $v_i = E_v(x_i)$ and $t_i = E_t(y_i)$ are the image and text features of sample $i$. The goal of vision-language contrastive learning is to maximize similarity for paired image-text examples while minimizing similarity for unpaired examples in the multimodal space. The image-to-text and the text-to-image contrastive losses are calculated by:
\begin{equation}
    {\cal L}^{i2t}_i=-\log\frac{\exp(S_{i,i}^{i2t} / \sigma)}{\sum_{j=1}^B \exp(S_{i,j}^{i2t} / \sigma)},
    \label{eq2}
\end{equation}
\begin{equation}
    {\cal L}^{t2i}_i=-\log\frac{\exp(S_{i,i}^{t2i} / \sigma)}{\sum_{j=1}^B \exp(S_{i,j}^{t2i} / \sigma)},
    \label{eq3}
\end{equation}
where $\sigma$ is a temperature parameter, $B$ represents the batch size. Then, the contrastive loss of sample $i$ is defined as:
\begin{equation}
    {\cal L}_i=\tfrac{1}{2}({\cal L}^{i2t}_i+{\cal L}^{t2i}_i).
    \label{eq4}
\end{equation}

\subsection{Label-Level Curriculum\label{curriculum1}}
The order of the pretraining data in the proposed cognitive orchestration is inspired by the cognitive processes of humans, which start with simpler aspects of a task and progressively tackle harder situations\cite{role, mind}. Existing public medical image datasets provide annotations primarily at two granularity: label-level and description-level. Label-level annotations\cite{jin2022fives, octid, irvin2019chexpert} typically provide a global diagnosis of the image, while description-level annotations\cite{keepfit,keepfitv2,johnson2019mimic} include the global diagnosis, as well as detailed descriptions of localized lesions and abnormalities. Obviously, deep neural networks face greater challenges in simultaneously integrating both global semantic information and localized lesion appearance\cite{spectral, fourier}. Therefore, we partition cognitive orchestration into two curricula according to the granularity of the supervisory signals associated with the images, as shown in \cref{fig2}.

Considering both clinicians and AI diagnostic models exhibit varying confidence when diagnosing different diseases from a single modality \cite{fundusreview}, we divide the label-level curriculum into three stages in decreasing order of modality sensitivity, as illustrated in \cref{fig2}(left). Specifically, differences in imaging purposes and physical principles across modalities cause each technique to emphasize certain morphological features, while other characteristics remain ambiguous or unobservable due to modality limitations. For example, OCT is designed to capture cross-sectional images of biological tissues. It relies on variations in light reflection from tissues at different depths. As a result, layer morphology is clearly visible in OCT, whereas information such as blood flow and color is difficult to visualize. Accordingly, in \textbf{Stage 1}, the classification is based on \textit{structural alterations that are visually identifiable and exhibit high modality-specificity}, such as hard exudates in CFP. In \textbf{Stage 2}, classification criteria are: \textit{multiple supporting signs and expert interpretation indicate a high-probability diagnosis ($>80\%$)}, examples include diabetic retinopathy in CFP or pneumonia in CXR. In \textbf{Stage 3} classification is assigned when \textit{reliable identification depends on complementary modalities, because current modality alone cannot provide definitive evidence or its manifestations are highly nonspecific and easily confounded with other pathologies}, such as glaucoma in CFP or fibrosis in CXR.

Multiple doctors are invited to categorize all diseases for each modality according to the stage-division criteria described above. In parallel, LLMs \cite{gpt,deepseek} are employed to perform the same division task, guided by the stage division criteria and their intrinsic diagnostic capabilities for specific diseases. All preliminary staging outcomes are reviewed and refined by a senior doctor. The final division of diagnostic sensitivity across all imaging modalities included in this study is presented in \cref{A.c1}.

\subsection{Description-Level Curriculum}
After acquiring global diagnostic ability through the label-level curriculum, the model further learns representation of local lesions and enhances its global diagnosis through the description-level curriculum. Given the absence of label information and the fact that textual descriptions can refer to multiple diseases, assigning each sample to a single category leads to misclassification. To address these challenges, we design a dedicated curriculum tailored to description-level data. Inspired by Lin \etal \cite{prototypes}, who demonstrates that performing pretraining from prototypes to general distributions can boost the performance of foundation models, we introduce a description-level curriculum that leverages intra-class samples representativeness. In medical images, due to the widespread presence of individual variations and comorbidities, samples farther from the class centroid are less influenced by these factors. As a result, their disease-related features are more distinct, making them more representative, As shown in \cref{fig1}(b), samples farther from the class centroid show typical AMD features such as hard exudates, while the nearer ones exhibit RPE tears that obscure these features. We require the model to first learn from these intra-class representative samples to capture basic disease features and then gradually recognize disease manifestations under increasingly complex conditions with individual variants or comorbidities, as shown in \cref{fig2}(right).

Specifically, we first extract image and text features from description-level data using a pretrained foundation model tailored to a single modality. We denote the resulting image and text features by $r_i^v$ and $r_i^t$, respectively. Meanwhile, we apply predefined prompt templates to all disease labels within the modality as mentioned in \cref{vlp} and encode them with the same text encoder, producing label embeddings $l_c$. We then compute the similarity between each sample's text feature and all label embeddings. The sample is assigned to the label with the highest similarity score:
\begin{equation}
    c=\arg\max(r_i^t \bm{l} ^T),
    \label{eq5}
\end{equation}
where $c$ is the cluster assigned to sample $i$, $\bm{l}$ represents the list of label embeddings $l_c$. After clustering, the centroid $u_c$ of each cluster is obtained by averaging the image features of the sample in that cluster. We then calculate the distance between all image feature $r_i^v$ and its cluster centroid $u_c$: $d(r_i^v, u_c) = \lVert r_i^v-u_c\rVert_2.$ However, the distribution of images exhibits a significant variation across diseases. Samples from certain diseases demonstrate a more dispersed distribution, while others form compact clusters. To address this heterogeneity, we normalize distances within each cluster by scaling them relative to the cluster radius $d_{max}$:
\begin{equation}
    d_i = d(r_i^v, u_c)/d_{max},
    \label{eq7}
\end{equation}
where $d_{max}$ represents the maximum distances within cluster $c$. All samples are then divided into $S$ stages in descending order of their distance from the centroid. During pretraining, the data belonging to the current stage will be added to the model training, where $S$ is a hyperparameter whose value depends on dataset scale, number of class, data distribution, etc. In the ablation experiment \cref{ab4}, we analyze the selection of $S$ for description-level curriculum.

\subsection{Self-Paced Asymmetric Contrastive Loss \label{SPCL}}
In medical images, discriminative features are often concentrated in small regions, while other regions are highly similar. This leads to substantial inter-class similarity among medical images. Most visual encoders in medical vision language foundation models are initialized with weights pretrained on natural images, where features are more macroscopic and less sensitive to localized variations. As a result, during the early stage of pretraining, the visual encoder will map semantically distinct medical images into similar feature representations, as shown in the top-left of \cref{fig1}(c). In contrast, the text encoder is typically initialized with weights from general-domain medical corpora, allowing it to effectively discriminate between descriptions of different diseases, as shown in the top-right of \cref{fig1}(c). Consequently, in early pretraining, the image-to-text alignment task is relatively straightforward due to the distinct and dispersed nature of text embeddings, whereas the text-to-image alignment task is more challenging owing to the tight distribution of image features.

Symmetric image-text contrastive loss \cref{eq4}, commonly used for natural images, is not suitable for medical vision-language contrastive pretraining, as it introduces gradient noise in the text-to-image alignment and leads to training imbalance that hinders effective multimodal convergence. To address this, we introduce a novel self-paced asymmetric image-text contrastive loss, which progressively increases the weight of the text-to-image part, thereby enabling a dynamic cognitive orchestration that transitions from easier to more difficult objectives during pretraining:
\begin{equation}
    {\mathcal L}_i=\tfrac{1}{2}({\mathcal L}^{i2t}_i+ \alpha(t,T){\mathcal L}^{t2i}_i).
    \label{eq7}
\end{equation}
where $t$ and $T$ represent current and total epoch, respectively, $\alpha(t,T)$ denotes the epoch-dependent weight of text-to-image contrastive loss. The weight schedule function can take any form. In our default setting, we adapt a simple linear schedule that gradually increases the weight from 0 to 1 throughout the entire pretraining process. For more details on the selection of different weight schedule functions formulation, please refer to \cref{ab2}. 


\begin{algorithm}[t]
\small
\setlength{\algomargin}{0pt}
\caption{Knowledge-driven cognitive orchestration for medical VLP}
\label{alg}
\textbf{Input: }{Pretraining dataset $D$, description-level data $D_d=\{(x_i,y_i)\}_{i=1}^N$, label set $L$, vision and text encoder $E_v$, $E_t$, pretrained vision and text encoder $E_v^{'}$, $E_t^{'}$, description-level curriculum stages $S$, epoch $T$}
\BlankLine
Generate \textbf{label-level curriculum} $[C_j]_{j=1}^3$

\tcp{Divide description-level data.}
Extract image and text features of all samples in $D_d$: $r_i^v=E_v^{'}(x_i),\quad r_i^t=E_t^{'}(y_i)$

Expand labels in $L$ using knowledge template $\pi(\cdot)$ and extract the list of label embeddings: $\bm{l}=E_t^{'}(\pi(L))$

Cluster each sample into a label $c$: $c=\arg\max(r_i^t \bm{l} ^T)$

Calculate image feature centroid: $u_c=avg(r_i^v),i\in c$

Compute normalized distance: $d_i = \lVert r_i^v-u_c\rVert_2/d_{max}^c$

\For{$s = 1,2,\dots,S$}{
    \If{$d_i\in [(s-1)/S,s/S)$}{
        Add sample $i$ into \textbf{description-level \\curriculum} $C_{3+s}$
    }
}
Union two level curricula: $C\leftarrow[C_j]_{j=1}^3 \oplus [C_{3+s}]_{s=1}^S$

\BlankLine
\For{$s = 1, 2, \dots, 3+S$}{
    \For{$t=1,2,\dots,T_s$}{
        Select $D_i\in C_s$ and randomly shuffle $D_i$
        
        Extract image and text features of samples in $D_i$: $v_i=E_v(D_i),\quad t_i=E_t(D_i)$

        Compute in-batch i2t and t2i contrastive losses using \cref{eq1}, \cref{eq2} and \cref{eq3}

        Compute the \textbf{weight of t2i loss}: $w=\alpha(t, T)$
        
        Compute loss: ${\mathcal L}_i=\tfrac{1}{2}({\mathcal L}^{i2t}_i+ w{\mathcal L}^{t2i}_i)$

        Optimize vision and text encoder $E_v$, $E_t$
    }
}

\textbf{Output:} {Final vision and text encoder $E_v$, $E_t$}
\end{algorithm}
\section{Experiments}
\begin{table*}[]
\centering
\renewcommand{\arraystretch}{1.05}
\caption{Result of zero-shot classification. The best result is \textbf{bolded} and the second-best is \underline{underlined}.}
\label{table1}
\resizebox{\textwidth}{!}{%
\begin{tabular}{ccc|cccc|cccc}
\Xhline{1pt}
Modality             & Dataset          & Metric & CLIP        & CL-log \cite{SPCL} & CL-logit \cite{CLfundus} & \textbf{MedKCO} & FILIP          & CL-log \cite{SPCL}  & CL-logit \cite{CLfundus} & \textbf{MedKCO} \\ \Xhline{1pt}
\multirow{3}{*}{CFP} & ODIR200×3        & ACC    & 0.772       & {\underline{0.835}}  & 0.567          & \textbf{0.863}     & 0.457          & {\underline{0.563}}          & 0.528           & \textbf{0.833}      \\
                     & REFUGE           & ACC    & {\underline{0.897}} & 0.861        & 0.803          & \textbf{0.947}     & \textbf{0.92}  & 0.875          & 0.832           & {\underline{0.891}}         \\
                     & FIVES            & AUC    & 0.676       & {\underline{0.682}}  & 0.659          & \textbf{0.729}     & 0.554          & \textbf{0.661} & 0.419           & {\underline{0.575}}         \\ \hline
\multirow{2}{*}{OCT} & OCTID            & ACC    & 0.709       & 0.688        & {\underline{0.743}}    & \textbf{0.778}     & {\underline{0.600}}    & {\underline{0.600}}    & {\underline{0.600}}     & \textbf{0.667}      \\
                     & OCTDL            & ACC    & 0.306       & {\underline{0.351}}  & 0.359          & \textbf{0.388}     & 0.307          & {\underline{0.322}}    & 0.292           & \textbf{0.420}      \\ \hline
\multirow{4}{*}{CXR} & CheXpert5*200    & ACC    & 0.384       & {\underline{0.466}}  & 0.443          & \textbf{0.526}     & 0.227          & {\underline{0.416}}    & 0.389           & \textbf{0.548}      \\
                     & RSNA-Pneumonia   & ACC    & {\underline{0.703}} & 0.527        & 0.643          & \textbf{0.714}     & \textbf{0.604} & {\underline{0.579}}    & 0.384           & 0.560               \\
                     & SIIM-Pneumothora & ACC    & 0.634       & {\underline{0.635}}  & 0.576          & \textbf{0.727}     & {\underline{0.635}}    & 0.565          & 0.507           & \textbf{0.762}      \\
                     & COVIDx           & ACC    & 0.463       & 0.353        & {\underline{0.469}}    & \textbf{0.564}     & {\underline{0.468}}    & 0.385          & 0.208           & \textbf{0.503}      \\ \hline
\multicolumn{3}{c|}{AVG}                         & {\underline{0.616}} & 0.600        & 0.585          & \textbf{0.693}     & 0.530          & {\underline{0.552}}    & 0.462           & \textbf{0.640}      \\ \Xhline{1pt}
\end{tabular}%
}
\end{table*}

\begin{table*}[]
\centering
\renewcommand{\arraystretch}{1.2}
\caption{Result of report generation. The best result is \textbf{bolded} and the second-best is \underline{underlined}.}
\label{table2}
\resizebox{\textwidth}{!}{%
\begin{tabular}{c|ccccccc|ccccccc|c}
\Xhline{1pt}
\multirow{2}{*}{Models} & \multicolumn{7}{c|}{OpenI}                                                                                           & \multicolumn{7}{c|}{MIMIC-CXR}                                                                                       & \multirow{2}{*}{AVG} \\ \cline{2-15}
                        & BLEU1          & BLEU2          & BLEU3          & BLEU4          & METEOR         & ROUGE          & CIDER          & BLEU1          & BLEU2          & BLEU3          & BLEU4          & METEOR         & ROUGE          & CIDER          &                      \\ \Xhline{1pt}
CLIP                & {\underline{0.362}}    & {\underline{0.216}}    & {\underline{0.146}}    & {\underline{0.105}}    & {\underline{0.140}}    & {\underline{0.311}}    & {\underline{0.293}}    & 0.259          & 0.157          & 0.102          & 0.072          & 0.112          & {\underline{0.246}}    & 0.104          & {\underline{0.188}}          \\
CL-logit \cite{SPCL}      & 0.359          & 0.211          & 0.140          & 0.098          & 0.138          & 0.309          & 0.255          & 0.258          & 0.154          & 0.099          & 0.070          & 0.111          & 0.242          & {\underline{0.107}}    & 0.182                \\
CL-log \cite{CLfundus}        & 0.328          & 0.197          & 0.134          & 0.097          & 0.132          & 0.303          & 0.286          & {\underline{0.269}}    & {\underline{0.162}}    & {\underline{0.105}}    & {\underline{0.074}}    & {\underline{0.114}}    & {\underline{0.246}}    & 0.106          & 0.182                \\
\textbf{MedKCO}  & \textbf{0.389} & \textbf{0.232} & \textbf{0.157} & \textbf{0.112} & \textbf{0.147} & \textbf{0.322} & \textbf{0.303} & \textbf{0.279} & \textbf{0.168} & \textbf{0.109} & \textbf{0.076} & \textbf{0.117} & \textbf{0.247} & \textbf{0.109} & \textbf{0.198}       \\ \hline
FILIP               & 0.378          & {\underline{0.232}}    & 0.158          & 0.114          & {\underline{0.149}}    & {\underline{0.326}}    & {\underline{0.316}}    & 0.252          & 0.151          & 0.099          & {\underline{0.070}}    & 0.109          & {\underline{0.242}}    & \textbf{0.097} & {\underline{0.192}}          \\
CL-logit \cite{SPCL}     & {\underline{0.380}}    & 0.230          & 0.156          & 0.113          & 0.147          & 0.321          & 0.290          & {\underline{0.264}}    & {\underline{0.159}}    & \textbf{0.104} & \textbf{0.072} & {\underline{0.112}}    & \textbf{0.245} & 0.091          & {\underline{0.192}}          \\
CL-log \cite{CLfundus}      & 0.379          & {\underline{0.232}}    & {\underline{0.159}}    & {\underline{0.115}}    & {\underline{0.149}}    & 0.324          & 0.311          & 0.254          & 0.152          & {\underline{0.100}}    & {\underline{0.070}}    & 0.106          & 0.241          & {\underline{0.093}}    & 0.145                \\
\textbf{MedKCO} & \textbf{0.394} & \textbf{0.239} & \textbf{0.163} & \textbf{0.118} & \textbf{0.151} & \textbf{0.328} & \textbf{0.330} & \textbf{0.270} & \textbf{0.162} & \textbf{0.104} & \textbf{0.072} & \textbf{0.113} & \textbf{0.245} & {\underline{0.093}}    & \textbf{0.199}       \\ \Xhline{1pt}
\end{tabular}%
}
\end{table*}

\subsection{Datasets}
We evaluated our method across three medical modalities: Color Fundus Photography (CFP), Optical Coherence Tomography (OCT) and Chest X-ray (CXR).\\
\textbf{CFP:} Following FLAIR \cite{flair} and KeepFIT\cite{keepfit}, we employed 31 public CFP datasets for model pretraining. These datasets integrate both label-level and description-level data. For evaluation, following KeepFIT, we assessed the model’s performance on ODIR$200\times3$ (\textbf{OOD}) \cite{flair}, FIVES \cite{jin2022fives} and REFUGE \cite{orlando2020refuge}.\\
\textbf{OCT:} Following KeepFIT v2 \cite{keepfitv2}, we utilized 11 label-level and one description-level datasets for pretraining and used OCTID \cite{octid} and OCTDL (\textbf{OOD})\cite{octdl} for evaluation.\\
\textbf{CXR:} Following MedCLIP \cite{medclip}, which is the first chest X-ray foundation model that simultaneously uses label and description-level data for pretraining. We utilized CheXpert \cite{irvin2019chexpert} as label-level data and MIMIC-CXR \cite{johnson2019mimic} as description-level data. Following GLoRIA \cite{huang2021gloria} and MedCLIP\cite{medclip}, we benchmarked the model on CheXpert5*200 \cite{huang2021gloria}, RSNA-Pneumonia \cite{rsna} and SIIM-pneumothorax. Moreover, we included COVIDx\cite{covidx} in our evaluation, which is an \textbf{OOD} dataset. For the zero-shot image-to-text retrieval and report generation task, we used OpenI\cite{openi} and MIMIC-CXR for evaluation. For more details regarding the datasets used in the training and evaluation phases, please refer to \cref{A.dataset}.

\subsection{Baselines}
We verified the effectiveness of MedKCO on two image-text contrastive pretraining frameworks: CLIP\cite{clip} and FILIP\cite{filip}. For both the CFP and the OCT modalities, all the data preprocessing and model architectures were consistent with those of FLAIR\cite{flair} and KeepFIT\cite{keepfit}. For the CXR modality, we referred to MedCLIP\cite{medclip}.

Additionally, we compared the proposed method with several curriculum learning methods. Self-paced learning \cite{SPL, SPCL} is a curriculum learning paradigm that dynamically adjusts the training process according to model feedback. In Jiang \etal \cite{SPCL}, the loss value of each training sample indicates its learning difficulty. A higher loss value indicates that the sample is relatively difficult and will initially be assigned a lower weight, which is gradually increased as training progresses. We refer to this strategy as \textbf{CL-log}. In Che \etal \cite{CLfundus}, the learning difficulty of each sample is determined by the model output logits. A lower logit indicates that the sample is relatively difficult and will initially be assigned a lower weight. We refer to this strategy as \textbf{CL-logit}. For more details of the pretraining and the baseline curriculum learning methods, please refer to \cref{A.baseline}.

\subsection{Implementation Details}
We used the CLIP baseline for each modality to extract image and text features for the description-level curriculum. We set $S=2$ for the description-level curriculum in all modalities. We randomly shuffled the data within each stage in the label and description-level curricula, and the data assigned to one stage did not appear in other stages.

The dimension of the projection head was set to 512 for every modality in CLIP and 256 for FILIP. The maximum text token length was set to 256. Data augmentation details are shown in \cref{A.detail}. A warm-up cosine scheduler was used during the first epoch. Pretraining was conducted on a single RTX A6000 GPU. All the compared models were trained with the same number of pretraining iterations.

\begin{figure*}[t]
\includegraphics[width=0.9\textwidth]{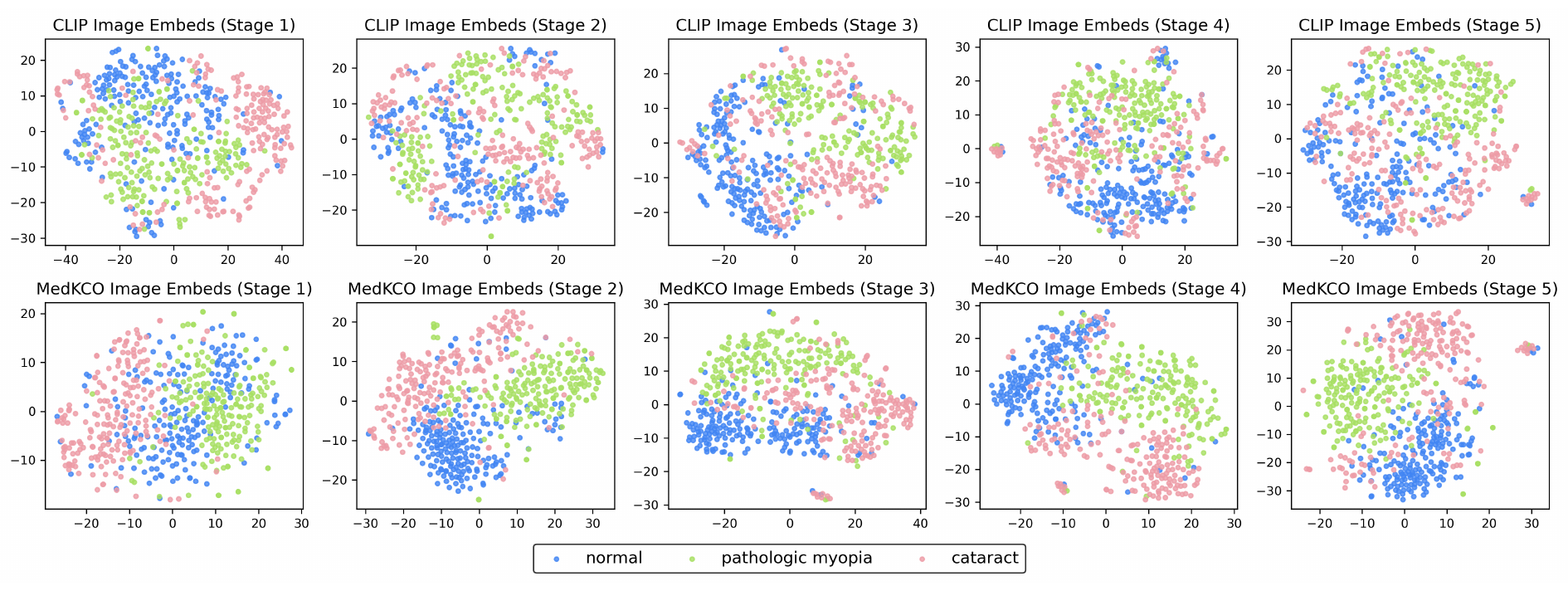}
\centering
\caption{Visualization of image feature at different stages on ODIR200×3 dataset under the CLIP framework.} \label{fig3}
\end{figure*}

\subsection{Main Results}
\textbf{Performance of Zero-Shot Classification.\label{ZS}}
To evaluate the performance of the proposed method, we conducted zero-shot classification experiments in three modalities. In which, ODIR$200\times3$, OCTDL and COVIDx served as the out-of-distribution (OOD) dataset for CFP, OCT and CXR modality, while the other datasets were under domain shift.
As summarized in \cref{table1}, the proposed method outperforms CLIP by 7.7\% and FILIP by 11\%, demonstrating its adaptability in different pretraining architectures. Moreover, it surpasses two curriculum learning baselines (CL-log and CL-logit) by 9.3\% and 10.8\% in CLIP framework, 8.8\% and 17.8\% in FILIP framework. These results highlight that cognitive orchestration guided by medical knowledge generates superior representation compared to conventional curriculum strategies. In particular, our method achieves the best results in all OOD datasets, confirming its robustness and effectiveness under distribution shift in medical zero-shot classification.\\
\textbf{Performance of Image-to-Text Retrieval Task.\label{Retrieval}} To evaluate the proposed method in vision-language tasks, we performed zero-shot image-to-text retrieval on the OpenI dataset and MIMIC-CXR test sets. As shown in \cref{table3}, the proposed method outperforms the baseline methods in all metrics and achieves an improvement of 1.7\%-5.5\% in CLIP framework and 2.4\%-3.8\% in FILIP framework on average. Especially in OpenI, a challenging OOD dataset, the proposed method achieves nearly twice the performance of the baseline. These results demonstrate that knowledge-driven cognitive orchestration facilitates the learning of more discriminative representations, thus enhancing retrieval performance.\\
\textbf{Performance of Report Generation.\label{report}} To further evaluate the transferability performance of the proposed model, we tested the model on the report generation task. We used the vision encoder of the pretrained foundation model to extract visual token sequences and adopt a Transformer encoder–decoder architecture to generate image captions. The training process was conducted on OpenI and MIMIC-CXR training set, and evaluated on each test set. More implementation details are available in \cref{A.RG}. As shown in \cref{table2}, the proposed method achieves the best results in all metrics, while the baseline curriculum learning methods fail to achieve any improvement. This indicates that the proposed knowledge-driven cognitive orchestration provides the foundation model with more transferable initial weights to downstream tasks.

\begin{table}[]
\centering
\renewcommand{\arraystretch}{1.1}
\caption{Result of zero-shot image-to-text retrieval. The best result is \textbf{bolded} and the second-best is \underline{underlined}.}
\label{table3}
\resizebox{\linewidth}{!}{%
\begin{tabular}{c|ccccccc}
\Xhline{1pt}
\multirow{2}{*}{Models}    & \multicolumn{3}{c}{OpenI}                  & \multicolumn{3}{c}{MIMIC-CXR}                & \multirow{2}{*}{AVG} \\ \cline{2-7}
                           & R@1          & R@5          & R@10         & R@1          & R@5           & R@10          &                      \\ \Xhline{1pt}
CLIP                 & 0.3          & 2.0          & 3.5          & 3.0          & 11.5          & 18.3          & 6.4                  \\
CL-logit \cite{SPCL}       & 0.6          & 2.1          & 3.6          & 3.4          & 11.9          & 18.5          & 6.7                  \\
CL-log \cite{CLfundus}        & {\underline{1.0}}    & {\underline{3.7}}    & {\underline{6.1}}    & {\underline{5.0}}    & {\underline{18.0}}    & {\underline{27.2}}    & {\underline{10.2}}           \\
\textbf{MedKCO}  & \textbf{1.2} & \textbf{4.0} & \textbf{6.4} & \textbf{6.4} & \textbf{21.7} & \textbf{31.7} & \textbf{11.9}        \\ \hline
FILIP                & 0.7          & 2.1          & 3.3          & {\underline{6.9}}    & 21.1          & 31.3          & 10.9                 \\
CL-logit \cite{SPCL}      & 0.7          & 2.2          & {\underline{4.0}}    & 5.3          & 17.9          & 27.9          & 9.7                  \\
CL-log \cite{CLfundus}        & {\underline{0.8}}    & {\underline{2.3}}    & 3.6          & 6.2          & {\underline{21.7}}    & {\underline{31.8}}    & {\underline{11.1}}           \\
\textbf{MedKCO} & \textbf{1.5} & \textbf{4.1} & \textbf{6.3} & \textbf{9.1} & \textbf{25.1} & \textbf{35.1} & \textbf{13.5}        \\ \Xhline{1pt}
\end{tabular}%
}
\end{table}

\subsection{Visualization of Cognitive Orchestration}
To better illustrate the effectiveness of knowledge-driven cognitive orchestration in enhancing representation learning, \cref{fig3} visualizes the image features extracted from the ODIR200×3 test set at different pretraining stages under the CLIP framework. As the curriculum progresses from stage 1 to stage 5, MedKCO produces a more structured and separable feature space compared to the baseline CLIP framework. This visually demonstrates how our method gradually enhances the discriminative capacity of the model by following a coherent cognitive learning order. More visualization of text features is provided in \cref{A.Visualization}.

\begin{figure*}[t]
\centering
\begin{minipage}[t]{0.58\textwidth}
    \centering
    \vspace{0pt} 
    \includegraphics[width=\textwidth]{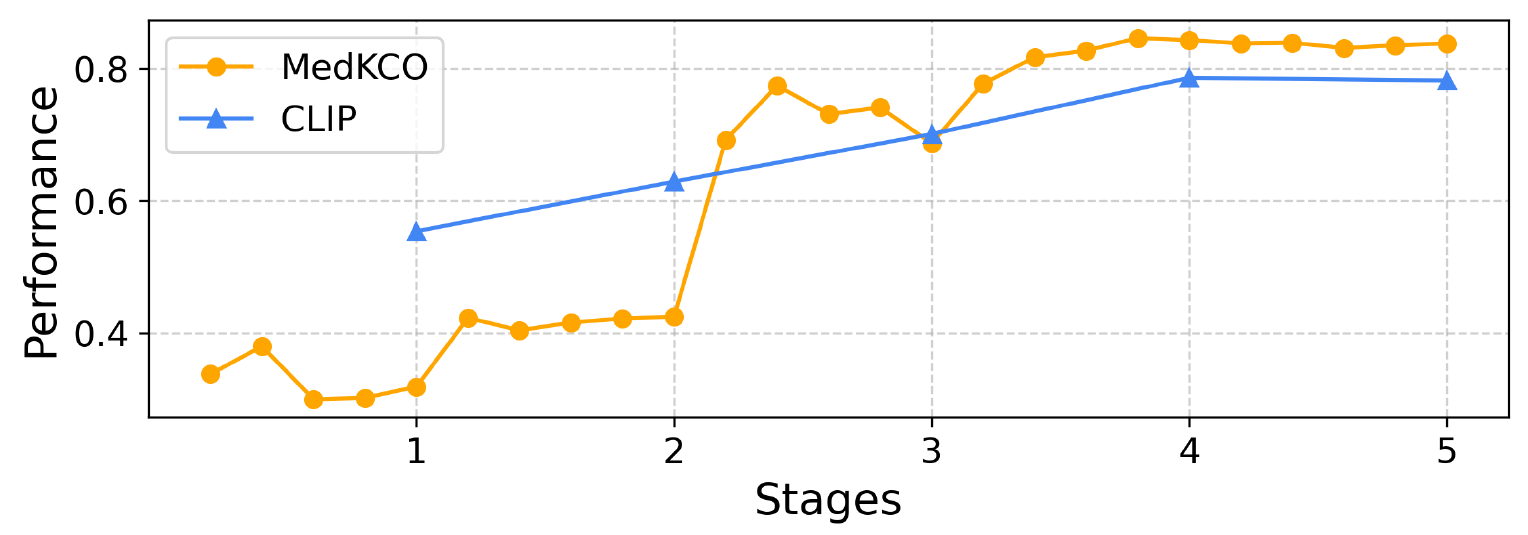}
    \caption{Efficiency of curriculum Learning. The proposed method has five epochs in each stage. For the baseline method, one stage represents one epoch.}
    \label{fig4}
\end{minipage}
\hspace{0.03\textwidth} 
\begin{minipage}[t]{0.37\textwidth}
    \centering
    \vspace{0pt} 
    \renewcommand{\arraystretch}{1.2}
    \captionof{table}{Comparison of weight schedule function.}
    \label{table4}
    \resizebox{\textwidth}{!}{%
    \begin{tabular}{ccccc}
    \Xhline{1pt}
    \multirow{2}{*}{Weight Function} & ODIR200×3 & FIVES & REFUGE & \multirow{2}{*}{AVG} \\ \cline{2-4}
                                     & ACC & ACC & AUC & \\ \Xhline{1pt}
    Segmented Linear & 0.810 & 0.694 & 0.910 & 0.805 \\
    Global Linear & \textbf{0.863} & \textbf{0.729} & \textbf{0.947} & \textbf{0.846} \\ \Xhline{1pt}
    \end{tabular}%
    }
    \vspace{0.5em}
    \captionof{table}{Comparison of stage division method.}
    \label{table5}
    \resizebox{\textwidth}{!}{%
    \begin{tabular}{ccccc}
    \Xhline{1pt}
    \multirow{2}{*}{Division Method} & ODIR200×3 & FIVES & REFUGE & \multirow{2}{*}{AVG} \\ \cline{2-4}
                                     & ACC & ACC & AUC & \\ \Xhline{1pt}
    Num of sample & 0.832 & 0.688 & 0.940 & 0.820 \\
    Distance & \textbf{0.863} & \textbf{0.729} & \textbf{0.947} & \textbf{0.846} \\ \Xhline{1pt}
    \end{tabular}%
    }
\end{minipage}
\end{figure*}
\begin{table*}[t]
\centering
\renewcommand{\arraystretch}{1.22}
\begin{minipage}[t]{0.62\textwidth}
\centering
\caption{Ablation study of different components on CFP modality using CLIP method.}
\label{table6}
\resizebox{\textwidth}{!}{%
\begin{tabular}{ccc|cccc}
\Xhline{1pt}
\multirow{2}{*}{Label-level Curriculum} & \multirow{2}{*}{\begin{tabular}[c]{@{}c@{}}Description-level \\ Curriculum\end{tabular}} & \multirow{2}{*}{\begin{tabular}[c]{@{}c@{}}Self-paced Asymmetric \\ Contrastive Loss\end{tabular}} & ODIR200×3 & FIVES & REFUGE & \multirow{2}{*}{AVG} \\ \cline{4-6}
                                      &                                                                                                   &                                                                                                    & ACC       & ACC   & AUC    &                      \\ \Xhline{1pt}
\ding{55}                                     & \ding{55}                                                                                                 & \ding{55}                                                                                                  & 0.772     & 0.676 & 0.897  & 0.782                \\
\ding{51}                                     & \ding{55}                                                                                                 & \ding{55}                                                                                                  & 0.847     & 0.704 & 0.908  & 0.820                \\
\ding{51}                                     & \ding{51}                                                                                                 & \ding{55}                                                                                                  & 0.890     & 0.658 & 0.969  & \underline{0.839}                \\
\ding{51}                                     & \ding{55}                                                                                                 & \ding{51}                                                                                                  & 0.827     & 0.746 & 0.934  & 0.836                \\
\ding{51}                                     & \ding{51}                                                                                                 & \ding{51}                                                                                                  & 0.863     & 0.729 & 0.947  & \textbf{0.846}       \\ \Xhline{1pt}
\end{tabular}%
}
\end{minipage}
\hfill
\begin{minipage}[t]{0.35\textwidth}
\centering
\renewcommand{\arraystretch}{1.2}
\caption{Number of stages in the description-level curriculum.}
\label{table7}
\resizebox{\textwidth}{!}{%
\begin{tabular}{ccccc}
\Xhline{1pt}
Num of & ODIR200×3      & FIVES          & REFUGE         & \multirow{2}{*}{AVG} \\ \cline{2-4}
Stages & ACC            & ACC            & AUC            &                      \\ \Xhline{1pt}
2      & \textbf{0.863} & \textbf{0.729} & \textbf{0.947} & \textbf{0.846}       \\
3      & 0.832          & 0.688          & \underline{0.940}    & 0.820                \\
4      & \underline{0.837}    & \underline{0.701}    & 0.923          & 0.820                \\ \Xhline{1pt}
\end{tabular}%
}
\end{minipage}

\end{table*}

\subsection{Ablation Study}
\textbf{Component Ablation.\label{ab1}} 
The pretraining method introduced in this paper consists of three core components: the label-level curriculum, the description-level curriculum, and the self-paced asymmetric contrastive loss. To assess the individual contribution of each module to the improvement of the baseline method, we perform ablation studies on the CFP modality under a zero-shot classification setting based on CLIP framework. The model equipped solely with the label-level curriculum was trained by first applying label-level curriculum, followed by training on description-level data without stage division.
As shown in \cref{table6}, label-level curriculum alone brings significant gains, especially in the case of OOD(ACC from 0.772 to 0.847), showing the importance of starting with easy-to-diagnose concepts. Adding the description-level curriculum further improves overall performance, indicating that fine-grained sample ordering within a category is beneficial to description-level data. The self-paced loss is especially crucial on dataset under domain shift, suggesting its effectiveness in handling high inter-class similarity. The combination of all three yields the best result, proving that data ordering and objective adjustment are complementary in cognitive orchestration.\\
\textbf{Weight Schedule Function in Self-Paced Asymmetric Contrastive Loss.\label{ab2}} 
As shown in \cref{eq7}, MedKCO implements a cognitive orchestration from easy to difficult in terms of pretraining objective by adaptively adjusting the weight of the text-to-image alignment task. The weight schedule function can take various forms, provided they can achieve a trend from low to high. In this work, we simply adopt a global linear schedule, where the weight linearly rises from 0 to 1 throughout training. We also compared this with a segmented linear schedule, which resets the weight to 0 at the start of each curriculum stage. Performance comparisons in \cref{table4} demonstrate that the global schedule yields superior results, as continuous weight accumulation enables smoother knowledge transfer across stages and alleviates the need for relearning.\\
\textbf{Stage Division in Description-Level Curriculum.\label{ab3}} 
In the description-level curriculum, stages are divided into equal-distance intervals based on the normalized distance between image features and their corresponding category feature centers. For comparison, we also evaluated a stage division strategy that equalizes the number of samples per stage. As shown in \cref{table5}, distance-based partitioning yields better performance, as the normalized distance reflects each sample’s representativeness within its category, enabling a clearer progression of cognitive difficulty. In contrast, number-based partitioning results in uneven difficulty across stages, hindering gradual cognition.\\
\textbf{Num of Stages in Description-Level Curriculum.\label{ab4}} 
The number of stages in the description-level curriculum is governed by several factors, such as dataset scale, number of class, data distribution and so on. To identify the optimal parameter, we performed a series of experiments, the results are summarized in \cref{table7}. We used this parameter throughout all of the experiments.

\subsection{Efficiency of Curriculum Learning \label{efficiency}}
Beyond improving performance, the curriculum learning paradigm also enhances training efficiency and reduces computational cost. \cref{fig4} compares the zero-shot classification performance of CLIP and MedKCO throughout training in CFP modality. In the early phase, MedKCO focuses on learning general knowledge and initially underperformed, as it has not yet encountered the test categories. With increasing task difficulty, the model gradually acquires more complex disease concepts, leading to steady performance gains. When description-level data are introduced, finer-grained textual supervision further enhances image–text alignment, yielding a notable performance boost that surpasses CLIP. Notably, the total number of training iterations remains the same for both methods.
\section{Conclusion}
In this work, we propose a knowledge-driven cognitive orchestration for medical VLP (MedKCO), which involves the ordering of pretraining data and the objective function. Experimental results show that MedKCO significantly improves downstream vision–language performance. By integrating domain knowledge, our method surpasses curriculum learning baselines, offering a cognitive-inspired paradigm for medical VLP. Future work will extend this framework to multimodal and automated curriculum.

\section{Acknowledgment}
This work is supported in part by the National Natural Science Foundation of China under Grant 62476054, 62576153, and the Fundamental Research Funds for the Central Universities of China. This research work is supported by the Big Data Computing Center of Southeast University.
{
    \small
    \bibliographystyle{ieeenat_fullname}
    \bibliography{main}
}

\clearpage
\setcounter{page}{1}
\maketitlesupplementary

\section{Pretraining and Evaluation Datasets \label{A.dataset}}
The proposed model and all baseline models were pretrained using both label-level and description-level data. The datasets used for each modality are summarized in \cref{table8}.

For the CFP modality, pretraining data span more than 96 fundus disease categories and include over 190k samples. For the OCT modality, the datasets cover more than 17 disease categories with over 180k samples. For the CXR modality, the pretraining datasets contained more than 14 disease categories and over 380k pretraining samples. During pretraining, we used only the MIMIC-CXR training set for the CXR modality.

For the zero-shot image-to-text retrieval task, we used the entire OpenI dataset for evaluation. For MIMIC-CXR, following CXR-CLIP \cite{cxrclip}, we evaluated performance solely on its test set. 

For the report generation task, we split the OpenI dataset into training, validation, and test sets by a ratio of 70\%, 15\% and 15\%, respectively. For the MIMIC-CXR dataset, we trained the model on its training set and evaluated the results on its test set.

\begin{table*}[]
\renewcommand{\arraystretch}{1.5}
\centering
\caption{Pretraining datasets of each modality.}
\label{table8}
\resizebox{\textwidth}{!}{%
\begin{tabular}{c|ll}
\Xhline{1pt}
Modality             & Label-level                                                                                                  & Description-level               \\ \Xhline{1pt}
\multirow{3}{*}{CFP} & EYEPACS, IDRID, RFMid, DEN, LAG, ODIR, PAPILA, PARAGUAY, STARE, ARIA, AGAR300, APTOS, FUND-OCT, JICHI,       & \multirow{3}{*}{MM-Retinal CFP} \\
                     & DiaRetDB1, DRIONS-DB, Drishti-GS1, E-ophta, G1020, HRF, ORIGA, ROC, BRSET, OIA-DDR, AIROGS, SUSTech-SYSU,    &                                 \\
                     & CHAKSU, DR1-2, Cataract, ScarDat                                                                             &                                 \\ \hline
\multirow{2}{*}{OCT} & RetinalOCT\_C8, Large\_Dataset\_of\_Labeled\_OCT, GAMMA1, STAGE1, STAGE2, glaucoma\_detection, GOALS, OIMHS, & \multirow{2}{*}{MM-Retinal OCT} \\
                     & OCTA\_500, DUKE\_DME, BIOMISA\_Retinal\_Image\_Database\_for\_Macular\_Disorders                             &                                 \\ \hline
CXR                  & CheXpert                                                                                                     & MIMIC-CXR                       \\ \Xhline{1pt}
\end{tabular}%
}
\end{table*}

\section{Detail of Label-level Curriculum \label{A.c1}}
When it comes to the label-level data of the three modalities used in this work, we first compiled all disease categories present in the datasets listed in \cref{A.dataset}, and then grouped these categories into three stages according to the classification criteria defined in \cref{curriculum1}. The class labels for each stage are presented in \cref{table9}.

\begin{table*}[]
\renewcommand{\arraystretch}{1.3}
\centering
\caption{The disease at different stage of each modality in label-level curriculum}
\label{table9}
\resizebox{\textwidth}{!}{%
\begin{tabular}{c|cl}
\Xhline{1pt}
Modality              & Stage                     & \multicolumn{1}{c}{Label}                                                                                       \\ \Xhline{1pt}
\multirow{20}{*}{CFP} & \multirow{4}{*}{Stage 1}  & hard exudates, soft exudates, microaneurysms, haemorrhages, media haze, drusens, tessellation, laser scar,      \\
                      &                           & optic disc cupping, tortuous vessels, asteroid hyalosis, optic disc pallor, exudates, cotton wool spots,        \\
                      &                           & colobomas, preretinal haemorrhage, myelinated nerve fibers, tilted disc, vitreous haemorrhage, large optic cup, \\
                      &                           & optic atrophy, fibrosis, silicon oil, scar, nevus, red small dots                                               \\ \cline{2-3} 
                      & \multirow{10}{*}{Stage 2} & no diabetic retinopathy, mild diabetic retinopathy, moderate diabetic retinopathy, severe diabetic retinopathy, \\
                      &                           & proliferative diabetic retinopathy, age-related macular degeneration, pathologic myopia, macular scar, shunt,   \\
                      &                           & branch retinal vein occlusion, epiretinal membrane, central retinal vein occlusion, optic disc edema,           \\
                      &                           & retinal traction, retinitis, retinal pigment epithelium changes,retinitis pigmentosa, haemorrhagic retinopathy, \\
                      &                           & central retinal artery occlusion, post traumatic choroidal rupture, choroidal folds, vasculitis, plaque,        \\
                      &                           & branch retinal artery occlusion, collaterals, maculopathy, severe hypertensive retinopathy, dragged disk,       \\
                      &                           & disc swelling and elevation, congenital disk abnormality, yellow-white spots flecks, abnormal macula,           \\
                      &                           & peripheral retinal degeneration and break, no proliferative diabetic retinopathy, hypertensive retinopathy,     \\
                      &                           & geographical age-related macular degeneration, abnormal optic disc, abnormal vessels, macular edema,            \\
                      &                           & increased cup disc, a disease, intraretinal microvascular abnormalities, retina detachment, normal              \\ \cline{2-3} 
                      & \multirow{6}{*}{Stage 3}  & diabetic macular edema, no referable diabetic macular edema, non clinically significant diabetic macular edema, \\
                      &                           & central serous retinopathy, anterior ischemic optic neuropathy, parafoveal telangiectasia, chorioretinitis,     \\
                      &                           & macular hole, optic disc pit maculopathy, haemorrhagic pigment epithelial detachment, Vogt-Koyanagi syndrome,   \\
                      &                           & glaucoma, Bietti crystalline dystrophy, neoplasm, no glaucoma, neovascular age-related macular degeneration,    \\
                      &                           & cataract, no cataract, macroaneurysm, cystoid macular edema, acute central serous retinopathy,                  \\
                      &                           & chronic central serous retinopathy, neovascularisation                                                          \\ \hline
\multirow{4}{*}{OCT}  & Stage 1                   & macular hole stage3, macular hole stage4, vitreomacular Interface Disease, epiretinal membrane                  \\ \cline{2-3} 
                      & \multirow{2}{*}{Stage 2}  & age related macular degeneration, drusen, diabetic macular edema, macular hole stage1,                          \\
                      &                           & macular hole stage2, normal, macular hole, central serous retinopathy, choroidal neovascularization             \\ \cline{2-3} 
                      & Stage 3                   & glaucoma, diabetic retinopathy, retinal artery occlusion, retinal vein occlusion                                \\ \hline
\multirow{4}{*}{CXR}  & Stage 1                   & Lung Opacity, Consolidation, Support Devices                                                                    \\ \cline{2-3} 
                      & \multirow{2}{*}{Stage 2}  & No Finding, Enlarged Cardiomediastinum, Cardiomegaly, Edema, Pneumonia, Atelectasis, Pneumothorax, Hernia,      \\
                      &                           & Pleural Effusion, Emphysema, Infiltration, Mass                                                                 \\ \cline{2-3} 
                      & Stage 3                   & Lung Lesion, Pleural Other, Fracture, Fibrosis, Nodule, Pleural\_Thickening                                     \\ \Xhline{1pt}
\end{tabular}%
}
\end{table*}

\section{Detail of the Baseline Models \label{A.baseline}}
For the \textbf{CFP} and \textbf{OCT} modality, following the data processing procedures and model architectures adopted in FLAIR \cite{flair} and KeepFIT \cite{keepfit}, we replaced categorical labels with descriptive phrases generated from domain knowledge templates to construct description-level supervision. In terms of model architecture, consistent with \cite{flair, keepfit}, we utilized ResNet50 as the vision encoder and Bio-ClinicalBERT as the text encoder. The models were pretrained using the CLIP InfoNCE loss function and FILIP fine-grained interactive language image pretraining loss. For the \textbf{CXR} modality, following MedCLIP \cite{medclip} and CXR-CLIP \cite{cxrclip}, we similarly replaced disease labels of varying confidence levels with descriptions generated from domain-knowledge templates. For the model architecture, we followed MedCLIP and CXR-CLIP by adopting ResNet50 as the vision encoder and Bio-ClinicalBERT as the text encoder. Pretraining used both the InfoNCE loss and the FILIP loss.

Additionally, we compared the proposed method with several curriculum learning baselines. Self-paced learning \cite{SPL, SPCL} is a training paradigm that dynamically adjusts the training process according to model feedback. In \cite{SPCL}, the loss value of each sample indicates its learning difficulty: samples with higher loss are treated as harder and are assigned lower weights in the early stages, with their weights increasing gradually as training progresses. This process is controlled by a loss threshold $\gamma$:
\begin{equation}
    \mathcal{L}=w_i {\cal L}_i(p, y) + \lambda(w_i) 
    \label{eq9}
\end{equation}

\begin{equation}
    w_i = \frac{1 + \exp(-\gamma)}{1 + \exp({\cal L}_i - \gamma)},
    \label{eq10}
\end{equation}

\begin{equation}
    \quad \mu _i = 1 + \exp(-\gamma) - w_i,
    \label{eq11}
\end{equation}

\begin{equation}
    \lambda(w_i)=\mu_iln(\mu_i) + w_i ln(w_i + \delta) - \gamma w_i,
    \label{eq12}
\end{equation}
where ${\cal L}_i(\cdot)$ represents the loss function for sample $i$, $p$ and $y$ correspond to the output logits and the target label, $\delta$ is a hyperparameter, which we set $1e^{-8}$ here. The threshold $\gamma$ was gradually increased by a fixed amount at each epoch. For vision-language contrastive pretraining under both the CLIP and FILIP frameworks, ${\cal L}_i(\cdot)$ corresponds to symmetric contrastive loss.

In \cite{CLfundus}, the learning difficulty of each sample is determined by the model’s output logits and a hyperparameter $\gamma$. A lower logit indicates that the sample is relatively difficult and that it will initially be assigned a lower weight. 
\begin{equation}
\mathcal{L} = - |p_t|^{\gamma}{\cal L}_i(p, y),
\label{eq13}
\end{equation}
where $p_t$ denotes the logit corresponding to the ground-truth class, with larger values indicating easier samples and smaller values indicating harder ones. The parameter $\gamma$ was adjusted to modulate the weight of easy and difficult samples throughout training, $p$ and $y$ correspond to the output logits and the target labels, respectively. In vision-language contrastive pretraining for the CLIP and FILIP frameworks, ${\cal L}_i(\cdot)$ refers to the symmetric contrastive loss.

\begin{figure*}[t]
\includegraphics[width=\textwidth]{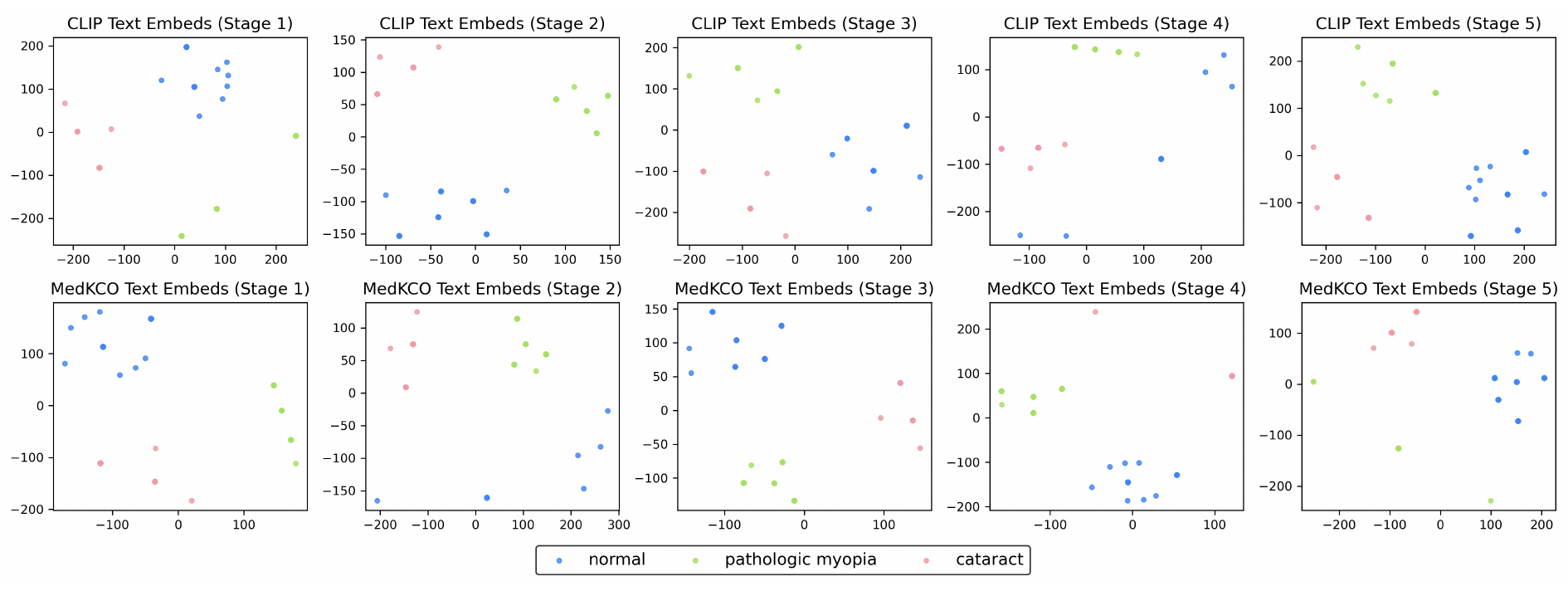}
\centering
\caption{Visualization of text feature at different stages on ODIR200×3 dataset under the CLIP framework.} \label{fig5}
\end{figure*}

\section{Implementation Detail \label{A.detail}}
\textbf{FILIP framework. }FILIP\cite{filip} computes fine-grained image–text and text–image similarities by leveraging sequences of local image and text features. The original FILIP adopts a ViT-based vision encoder. In this paper, we replaced it with ResNet50 for consistency with FLAIR, KeepFit, KeepFit v2, Med-CLIP and CXR-CLIP. Specifically, in the output part of the vision encoder, we removed the global average pooling layer and the corresponding normalization layer, obtaining a feature map of size $B\times H\times W\times C$. The tensor was then flattened into a feature sequence of size $B\times (HW)\times C$.\\
\textbf{Data augmentation. }For CFP and OCT modality, following the design of FLAIR \cite{flair} and KeepFIT \cite{keepfit} series, we resized the image to $512 \times 512$. During pretraining, we applied random image augmentations including horizontal flips, random rotations of $[-5, 5]$ degrees, zoom scaling sampled from $[0.9, 1.1]$, and color jitter. We used AdamW with a base learning rate of $1e^{-4}$ and weight decay of $1e^{-5}$. For the CXR modality, following MedCLIP \cite{medclip} and CXR-CLIP \cite{cxrclip}, we resized the image to $256 \times 256$. Image augmentations were applied using horizontal flipping with 0.5 probability; color jittering with brightness and contrast ratios of $[0.8; 1.2]$; random affine transformation with degree sampled from $[-10; 10]$, maximum translation rate 0.0625 and scale factor in $[0.8; 1.1]$. We used AdamW with a base learning rate of $5e^{-5}$ and a weight decay of $1e^{-4}$.

\section{Report Generation \label{A.RG}}
For the report generation task, we used the vision encoder of the pretrained foundation model to obtain a sequence of visual features. Specifically, with the ResNet50 architecture used in this paper, we extracted the feature maps before the final average pooling layer and flatten their spatial dimensions. The resulting sequence of visual features was first processed by a Transformer encoder to form visual memory. Then, we concatenated the text tokens with this visual memory and fed them into the Transformer decoder. The training was conducted under an auto-regressive manner.

\section{Visualization of the Text Embedding\label{A.Visualization}}
Since the text encoders of medical VLPs are typically initialized with models pretrained on large-scale medical corpora, they inherently possess strong semantic modeling capabilities across a wide range of diseases. Consequently, textual features tend to exhibit a relatively sparse distribution in the early stage of VLP pretraining. As shown in \cref{fig5}, both MedKCO and CLIP demonstrated strong disease modeling abilities through their text encoders at different stages of pretraining. Meanwhile, \cref{fig3} shows that MedKCO achieved a more powerful visual representation modeling performance. Together, these results indicate that the proposed knowledge–driven cognitive orchestration method effectively enhances the ability of medical VLP to model disease-related image features.

\end{document}